\documentclass[letterpaper, 10 pt, conference]{ieeeconf}  

\usepackage[T1]{fontenc}
\usepackage{adjustbox} 
\usepackage{multirow}
\usepackage{tabularx}
\usepackage{array}
\usepackage{booktabs}
\usepackage{amssymb}

\usepackage{hyperref}       
\usepackage{url}

\IEEEoverridecommandlockouts                   
\title{\LARGE \bf
NaturalVLM: Leveraging Fine-grained Natural Language for Affordance-Guided Visual Manipulation}

\author{Ran Xu$^{*}$, Yan Shen$^{*}$, Xiaoqi Li, Ruihai Wu, Hao Dong
\thanks{All authors are with School of CS, Peking University and National Key Laboratory for Multimedia Information Processing.
Xiaoqi Li is also with Beijing Academy of Artificial Intelligence (BAAI).
}
\thanks{
* Equal Contribution. Author ordering determined alphabetically.
}
\thanks{
Corresponding to hao.dong@pku.edu.cn}
}

\begin{document}

\maketitle
\thispagestyle{empty}
\pagestyle{empty}

\begin{abstract}

Enabling home-assistant robots to perceive and manipulate a diverse range of 3D objects based on human language instructions is a pivotal challenge.
Prior research has predominantly focused on simplistic and task-oriented instructions, \emph{i.e.}, "Slide the top drawer open". 
However, many real-world tasks demand intricate multi-step reasoning, and without human instructions, these will become extremely difficult for robot manipulation.
To address these challenges, we introduce a comprehensive benchmark, NrVLM, comprising 15 distinct manipulation tasks, containing over 4500 episodes meticulously annotated with fine-grained language instructions.
We split the long-term task process into several steps, with each step having a natural language instruction. 
Moreover, we propose a novel learning framework that completes the manipulation task step-by-step according to the fine-grained instructions.
Specifically, we first identify the instruction to execute, taking into account visual observations and the end-effector's current state.
Subsequently, our approach facilitates explicit learning through action-prompts and perception-prompts to promote manipulation-aware cross-modality alignment.
Leveraging both visual observations and linguistic guidance, our model outputs a sequence of actionable predictions for manipulation, including contact points and end-effector poses.
We evaluate our method and baselines using the proposed benchmark NrVLM. The experimental results demonstrate the effectiveness of our approach.
For additional details, please refer to \href{https://sites.google.com/view/naturalvlm}
{https://sites.google.com/view/naturalvlm}.

\end{abstract}

\section{INTRODUCTION}

Language serves as a crucial means for robots to engage with the world~\cite{jang2022bc,shridhar2023perceiver,srivastava2022behavior}. 
Robots must not only comprehend language instructions but also integrate them with real-time visual observations to make informed predictions for manipulation tasks.
The existing benchmark VLMbench~\cite{zheng2022vlmbench}, provides high-level language instructions to guide robot agents, such as ``pick up the red plate``. However, relying solely on high-level instructions presents challenges, particularly for complex or unfamiliar tasks. Without the inclusion of low-level language instructions for guiding robots through each step of a task, successful task completion becomes exceedingly difficult.
While some previous efforts have harnessed large language models~\cite{huang2023voxposer,jin2023alphablock} to generate low-level instructions, these instructions tend to lack diversity in language style and often fail to accurately describe end-effector commands. This limitation arises because these models were not specifically trained for manipulation tasks. The detailed illustration is shown in Sec.~\ref{Ins Generation}.

\begin{figure}[t]
  \centering
  \includegraphics[scale=0.33]{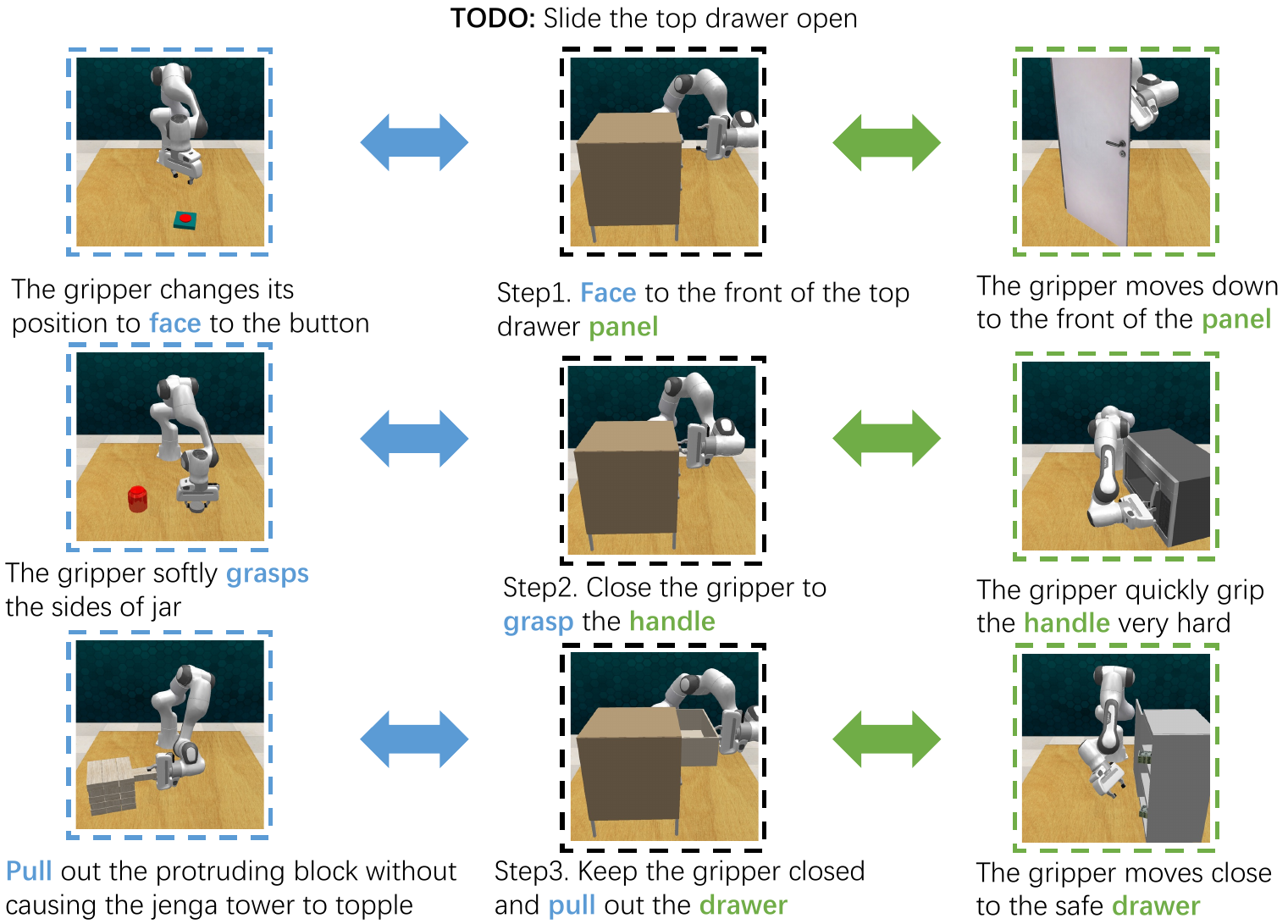}
  \caption{Illustration on the fine-grained instructions. The leftmost and rightmost pairs represent action-prompt and perception-prompt bases respectively. In the center column are the manipulation steps for "Slide the top drawer open," each accompanied by fine-grained language instructions. If the current task's manipulation step shares the same action or noun phrase as another task's manipulation step in the fine-grained language instruction, cross-modal alignment will be conducted using the features of the action-prompt base and the perception-prompt base.}
  \label{fig:teaser}
\end{figure}

In this work, we aim to address existing limitations by introducing a novel task: low-level visual language manipulation. This task involves employing natural language to provide step-by-step instructions for visual manipulation tasks. To enable this, we propose the NrVLM dataset, which offers a collection of natural language instructions meticulously paired with manipulation episodes collected from V-REP simulation. Our dataset comprises an extensive dataset of 4500 episodes featuring 82 distinct variations across 15 tasks.
We split the robot's manipulation task into discrete steps, each thoughtfully annotated in natural language. 
In Fig.\ref{fig:teaser}, we illustrate the fine-grained instructions for ``Slide the top drawer open''. These instructions encompass crucial details, \emph{i.e.}, the required robot actions, designated objects for interaction, and the necessary end-effector state for each step.
The purpose of these language instructions is to guide the agent in successfully executing manipulation tasks.
To the best of our knowledge, the NrVLM dataset we present stands as a pioneering benchmark, uniquely combining manipulation trajectories with free-form instructional language, offering an invaluable resource for advancing research in this field.

Along with the benchmark, we further devise a framework that enables the agent to follow the instructions and execute step by step. 
This framework effectively utilizes a multi-modal approach, incorporating various information sources including visual observations of the current scene, the present state of the end-effector, high-level language instructions, and fine-grained language instructions to generate actions for each individual step.
Concretely, when presented with a sequence of low-level instructions, our model initially assesses the required execution step by analyzing the current visual scene, the end-effector status, and the high-level linguistic instruction.
Subsequently, in order to fully exploit information from multiple modalities and predict reliable manipulation actions, we facilitate explicit learning through manipulation-aware cross-modality feature alignment as shown in Fig.~\ref{fig:teaser}.
To enhance this alignment process, we establish both action-prompt and perception-prompt bases in advance since the features associated with actions and objects remain consistent across different tasks. 
The action-prompt base pairs action phrases (e.g., "pull", "grasp") with their corresponding action features, while the perception-prompt base pairs noun phrases (e.g., "drawer", "handle") with their corresponding object features.
These pre-established priors significantly aid in aligning the respective multi-modal features, resulting in better predictions for manipulation actions.

To investigate the difficulty of the benchmark NrVLM and to evaluate the performance of our proposed method, we conduct extensive evaluations with four competitive visual language approaches and an ablated version. The results demonstrate the effectiveness of our framework against other methods.

In summary, we make the following contributions in this work:
\begin{itemize}
    \item We present the NrVLM, a comprehensive benchmark that combines diverse manipulation trajectories with fine-grained natural instructions, facilitating the agents in executing complex tasks sequentially.
    \item We propose a novel framework that enables the agent to utilize fine-grained instructions and acquire the manipulation-aware multi-modality alignment.
    \item Experimental results against four baselines validate the effectiveness and superiority of the proposed approach.

\end{itemize}

\section{RELATED WORK}
\subsection{Robotic Manipulation Benchmarks}
There are plenty of benchmarks related to visual-language robotic tasks~\cite{yu2020meta,ehsani2021manipulathor,shridhar2020alfred,james2020rlbench,zheng2022vlmbench}.
ALFRED~\cite{shridhar2020alfred} is proposed for vision-and-language navigation and virtual object rearrangement tasks between different room-scale locations.
RLBench~\cite{james2020rlbench} provides a benchmark and learning environment for both ‘robot learning’ and ‘traditional’ methods. It is a large-scale benchmark consisting of 100 completely unique, hand-designed tasks, which are diverse in task, variation, and episodes.
Following RLBench, VLMBench~\cite{zheng2022vlmbench} collects a robot manipulation benchmark on 3D tasks with visual observation and compositional language instructions. It categorizes manipulation tasks into various meta manipulation actions
according to the constraints of robot trajectories for the first time. 
Different from previous work, our NrVLM benchmark offers low-level natural language instructions that annotate the robotic manipulation trajectory step by step. Such instructions aim to guide the agent to accomplish long-term and intricate tasks.

\subsection{Vision-and-Language Manipulation}
Among visual language manipulation approaches~\cite{zhang2021invigorate,shridhar2022cliport, jang2022bc, zheng2022vlmbench, huang2023instruct2act, jiang2022vima}, BC-Z~\cite{jang2022bc} develops imitation learning system to enable a vision-based robotic manipulation system. It aims to generalize to novel tasks and address a long-standing challenge in robot learning. 
Furthermore, PERACT~\cite{shridhar2023perceiver} is an innovative behavior-cloning agent that utilizes a Perceiver Transformer~\cite{jaegle2021perceiver} to encode both language and voxel scenes. The model exhibits strong performance in multi-task manipulation.
In contrast, our proposed framework focuses on learning to follow low-level instructions to carry out step-by-step manipulation actions. Our approach adopts an action-prompt and perception-prompt based system that enables a thorough, manipulation-aware understanding of multiple modalities.

\section{Fine-grained Instructed Manipulation}

\subsection{Problem definition}
In the low-level visual language manipulation task, the agent is provided with fine-grained language instructions to follow in order to complete the manipulation task. 
Specifically, to begin with, the agent is provided with a sequence of natural language instructions $\mathcal{L}=\{l_{1},l_{2},,,,l_{n}\}$ (n denotes the maximum length of the language instructions) and current visual observations, including multi-view RGB images, depth images, and segmentation information.
Given the initial state of the robot, the agent needs to predict an executable action command for the robot based on linguistic and visual information.
After executing the action, the agent will obtain a new set of visual observations and continue with predicting more actions until it completes the sequence of low-level instructions.

\begin{figure}[t]
  \centering
  \vspace{5mm}
  \includegraphics[scale=0.37]{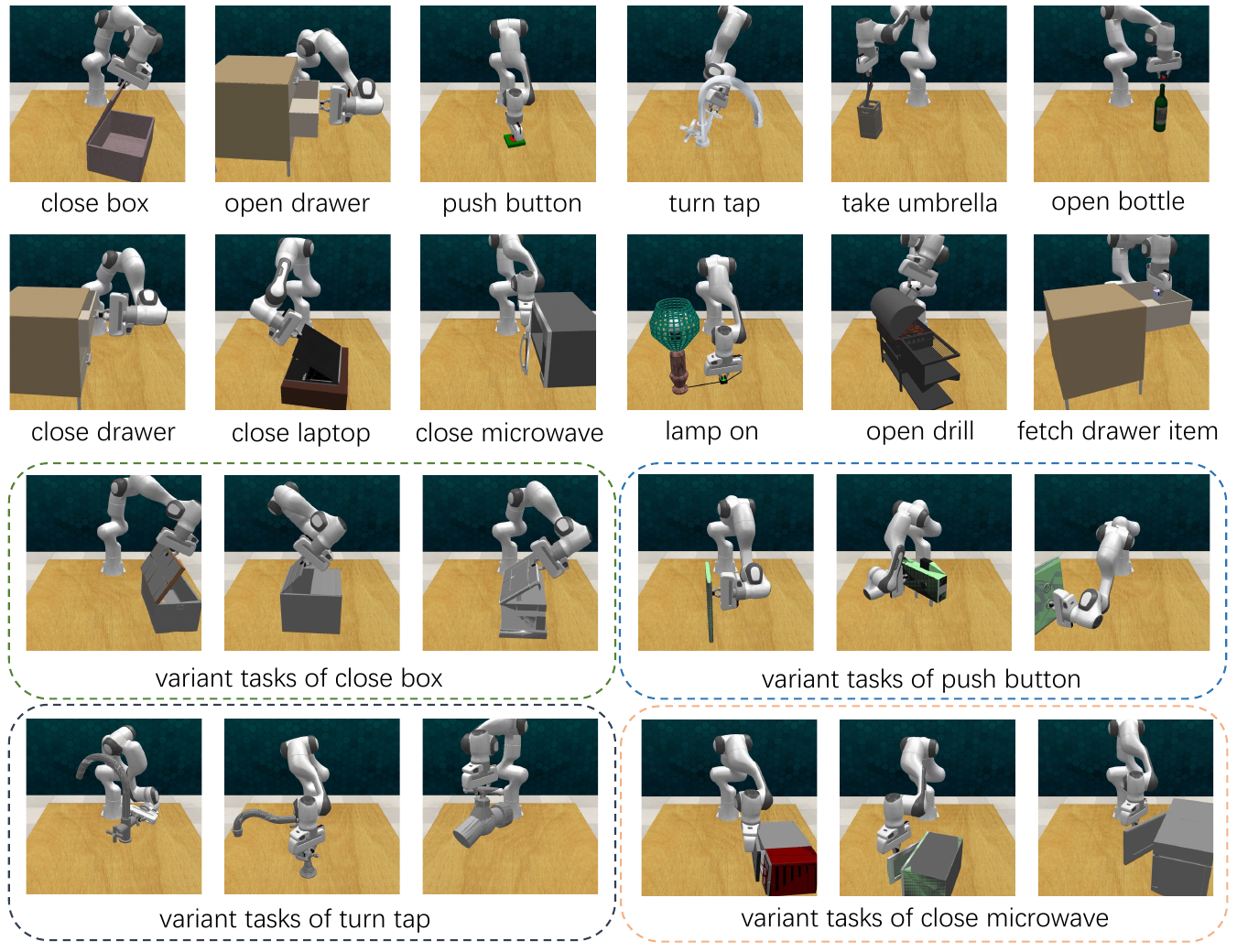}
  \caption{We introduce NrVLM, a comprehensive benchmark comprising multiple manipulation tasks annotated with fine-grained natural language instructions. Visualization of select tasks from the benchmark is presented in the top two rows. Additionally, we introduce difference task variations to enrich the diversity and complexity of the benchmark, as demonstrated in the bottom two rows.}
  \label{fig:benchmark}
\end{figure}

\subsection{Data Collection}
Drawing inspiration from RLBench \cite{james2020rlbench}, we conduct our own data collection for manipulation episodes across 15 distinct tasks. These manipulation tasks encompass a spectrum from simple tasks (\emph{e.g.}, reach target) to more challenging ones (\emph{e.g.}, stack cups), with the number of manipulation steps required ranging from 1 to 20. Figure~\ref{fig:benchmark} visually showcases some of these tasks included in our benchmark.
This diverse array of task difficulties facilitates a comprehensive evaluation of the trained agents' performance.

To enhance the richness and complexity of our dataset, we introduce variations within each task. These variations involve introducing different object instances or altering shape geometries, thereby augmenting the diversity of challenges. 
Specifically, for manipulation tasks like "close-box" or "close-microwave", we select specific object instances from PartNet-Mobility dataset~\cite{mo2019partnet} that can replace the original objects provided in RLBench. For example, in the case of the task “close-microwave”, the size, type, geometry, and color of the manipulated microwave vary in different task variations. These selected objects exhibit different properties and characteristics, contributing valuable variations to our dataset. We illustrate some of these task variations in the two bottom rows of Figure~\ref{fig:benchmark}.
Each task typically offers between 1 to 15 variations.

From these variations, we can generate an infinite number of manipulation episodes for training and evaluation.
Within each variation, manipulation episodes differ in various aspects, such as the initial position and orientation of objects on the workbench, and the objects' relative spatial arrangements. Consequently, this results in diverse manipulation trajectories. This design enriches the diversity of manipulation trajectories required to accomplish each task, thereby bolstering the robustness of the trained agent.

All these tasks are categorized into three sets: a training set, a validation set, and a test set. The training set includes a total of 8 manipulation tasks with 46 variations, and both the validation and test sets include a total of 15 manipulation tasks (8 training tasks with an extra 7 novel tasks) with 82 variations. Each task comprises 300 manipulation episodes. The allocation of episodes between the training, validation, and test sets follows a ratio of 10:1:1. Our objective with this setup is to assess the agent's capability to successfully perform these novel tasks with the guidance of low-level natural language instructions. This evaluation helps us understand the agent's adaptability and generalization to new tasks and scenarios.

\subsection{Fine-grained Instruction Annotation}
In previous research on Visual Language Manipulation (VLM) tasks, language instructions primarily consisted of high-level descriptions. For example, an instruction for the "close-box" task might be as simple as "shut the lid of the box." The high-level descriptions remain totally the same for different manipulation attempts in the same task.
While these high-level instructions convey the overall task objective, we have observed that most manipulation tasks should be accomplished by several steps and are naturally composed of several key actions. Therefore, these high-level instructions lack explicit guidance for the robot on how to perform these actions and accomplish the task step by step. To address this gap, we introduce the concept of fine-grained language instructions to guide the agent to complete the task step by step. Besides, the fine-grained language instructions are more precise and diverse compared to the unchanged high-level descriptions.

To generate fine-grained language instructions according to the sequential manipulation steps, we first split each expert manipulation demonstration into several slices. Following previous work~\cite{johns2021coarse, shridhar2023perceiver}, we conducted keyframe action extraction to split the manipulation episode. The principle of keyframe action extraction is based on two criteria: (1) the velocity of the robot joints approaches zero, and (2) the gripper's open state changes. After splitting the demonstrations according to the keyframe actions, annotators can describe the precise robotic manipulation and provide fine-grained instructions for each step.

These annotated natural language instructions include action verbs, object noun phrases, and other details that explicitly outline how the end-effector should complete each step of the task.
It is worth noting that the number of annotated natural instructions is closely related to the complexity of the task. Tasks of greater complexity, which demand a larger number of steps for completion, are accompanied by longer sets of fine-grained instructions.
To maintain diversity and prevent the emergence of overly uniform linguistic styles in fine-grained instructions across different manipulation episodes, each task requires the involvement of at least ten distinct annotators to produce the fine-grained natural instructions. Each annotator undergoes an independent annotation process without access to others' results.
This approach significantly enhances the diversity of language instructions, ensuring a broader range of language styles and expressions.

\begin{figure*}[t]
\vspace{3mm}
  \centering
  \includegraphics[scale=0.56]{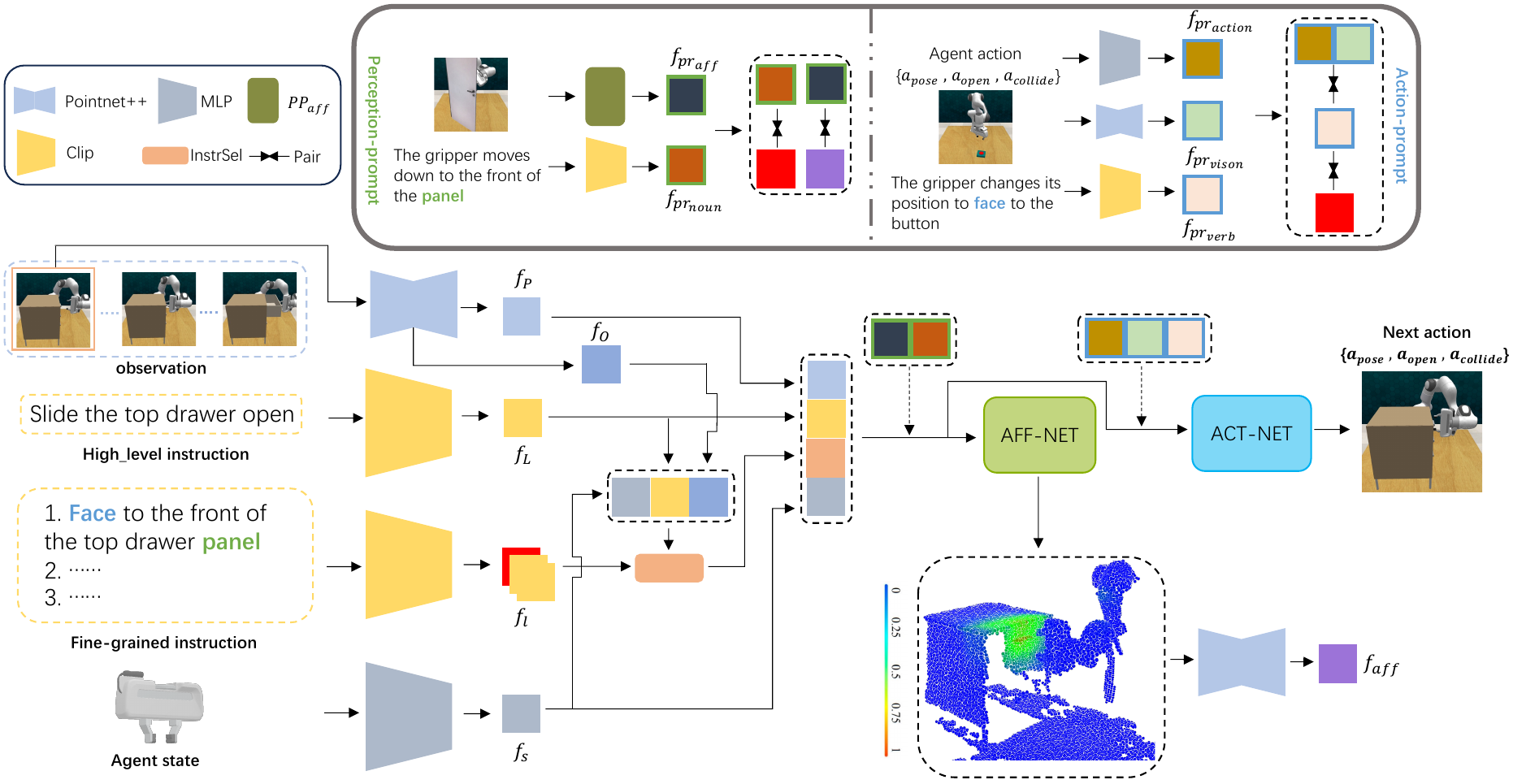}
  \caption{\textbf{The overall framework.} The bottom part shows the manipulation process, where the Instruction Selection network (InstrSel) selects the appropriate fine-grained language instruction, the Affordance network (AFF-NET) predicts the object-centric affordance map, and the Actor network (ACT-NET) predicts the gripper action.
  The top part shows the alternative perception-prompt module and action-prompt modules, they enhance the Affordance and Actor networks by aligning the noun-related perception-prompt set and verb-related action-prompt set. The two dotted arrows before Affordance and Actor networks indicate that the prompt modules are optional.
  The entire method is trained in an end-to-end manner.}
  \label{framework}
    \vspace{-3mm}
\end{figure*}

\section{METHOD}

\subsection{Task Formulation}
In our tasks, each demonstration is split into several steps, with each step paired with a fine-grained language instruction $l$ and a keyframe action $a$. 
At each step, the agent receives visual observations, linguistic guidance, and agent state.
The visual observation is the scene point cloud acquired from the RGB-D taken by depth cameras, while the linguistic information contains both high-level and fine-grained language instructions. The agent state is a 4-DoF vector that includes 1-DoF gripper open state and 3-DoF gripper position.
Given these multi-modal information, the agent is expected to predict the gripper action $a = \left\{a_{pose}, a_{open}, a_{collide}\right\}$, which includes a 6-DoF pose, a 1-DoF gripper open state, and a 1-DoF collision state that determines whether the motion-planner uses collision avoidance to reach an intermediate pose.
By formulating this manipulation problem as a behavior-cloning problem, we adopt key action as expert action to supervise the agent's predicted action for each step, enabling the agent to learn from the expert demonstrations.

\subsection{Framework Overview}
Fig.~\ref{framework} presents an overview of our proposed framework. 
As shown in the bottom part, at each step, given the current point cloud observation, agent state, high-level instruction, and a sequence of fine-grained language, Instruction Selection (InstrSel) network aims to select the appropriate fine-grained language instruction from a sequence of instructions.
Next, the Affordance network (AFF-NET) establishes manipulation-aware multi-modality alignment and predicts the object-centric affordance map.
The affordance map indicates where the agent should act (e.g., to open the door, the affordance map highlights the door handle and indicates where to grasp).
Based on the affordance map, we filter out low-rated regions and select a contact point.
We further incorporate Actor network (ACT-NET) to predict the position offset based on the selected contact point, along with gripper rotation, open state, and collision state.
The offset is introduced since different primitives require varying offsets in different directions at the contact point. For instance, in the case of lifting, the gripper needs to be positioned directly above the contact point vertically.

In the top part of the figure, we aim to bridge the three modalities including vision, language, and manipulation.
We introduce two prompt modules: the perception-prompt module and the action-prompt module. The perception-prompt module enhances the Affordance network by aligning the noun-related perception-prompt set, while the action-prompt module improves the Actor network by aligning the verb-related action-prompt set.
For example, in the instruction "grasp the drawer handle", the noun "handle" refers to a specific object part that the agent should interact with, while the verb "grasp" indicates the action should take. Thus, the instruction verb connects the language and manipulation modalities, while the noun connects the language and vision modalities.

\subsection{Instruction Selection}
Fine-grained language instructions provide the agent with a step-by-step understanding of the manipulation task. After receiving the sequence of fine-grained language instructions, the agent first selects the appropriate instruction based on the current scene. To process the multi-modal information, we use pre-trained CLIP~\cite{clip} to extract the text feature $f_{l_i}$ for each instruction and PointNet++~\cite{qi2017pointnet++} to extract the global feature of the entire point cloud $f_o$. We also use the pre-trained CLIP to extract the high-level instruction feature $f_L$, and an MLP network to encode agent state feature $f_s$.
Subsequently, in \textit{Instruction Selection network}, we calculate the similarity of each fine-grained language instruction $f_{l_i}$ with the current scene information including $f_o$, $f_L$, and $f_s$.
We then apply the softmax operation on the similarity score
and obtain the normalized weight to select the most related instruction. By summing up the fine-grained language features $f_{l_i}$ according to the normalized weight, we can obtain the weighted feature of the fine-grained language instruction $f_{l}$.
Normalized weight is formulated as one-hot vector, which is supervised by ground-truth one-hot vector under NLL loss.
This approach enables the agent to select the most related instruction by understanding the fine-grained instruction and the current scene. 

\subsection{Affordance Network and Actor Network}
The \textit{Affordance network} predicts affordance map for manipulation task based on multi-modal information. The map indicates the manipulation region for the agent.   
Specifically, we adopt the PointNet++~\cite{qi2017pointnet++} to extract the per-point feature $f_p$. Note that we use the same PointNet++ encoder for both extracting the per-point feature $f_p$ and the global feature $f_o$, while using the segmentation decoder and the classification decoder to decode the feature separately. The Affordance Network is implemented as an MLP architecture, which takes in the feature concatenation of $f_p$, $f_s$, $f_l$, $f_L$,
and outputs the affordance score $\in [0, 1]$ for each point. Point-wise affordance score forms the affordance map, from which we select a contact point $p' \in \mathbb{R}^{3}$ with a high affordance score.  
Following ~\cite{weng2022fabricflownet}, we acquire the ground-truth affordance using the recorded action in the expert demonstration, by placing a 3D Gaussian on the ground-truth interaction location. We then supervise the Affordance network by using the binary cross-entropy (BCE) loss between the predicted affordance map and the ground-truth.

The \textit{Actor network's} predicts the action based on the selected contact point. 
For different primitive types, such as lifting or pulling open, the action needs to be carried out with varying directional offsets based on the contact point to accomplish the task.
The Actor network thus predicts the movement $a_{move} \in \mathbb{R}^3$ based on the selected contact point $p'$. The gripper position is then calculated using the formula $a_{position} = p' + a_{move}$. 
Additionally, it also predicts the rotation $a_{rot}$ in a quaternion format, the gripper open state $a_{open}$, and the collision state $a_{collide}$. 
The predicted gripper position $a_{position}$ and rotation $a_{rot}$ are combined to form the gripper pose $a_{pose}$.
The Actor network is implemented both encoder and decoder in MLP architecture. 
The encoder takes the feature concatenation of $f_{p'}$, $f_s$, $f_l$, $f_L$,
and the four decoders predict $a_{move}$, $a_{rot}$, $a_{open}$, and $a_{collide}$ respectively.
To supervise the predicted gripper action, we use L1 loss for $a_{move}$, quaternion distance loss for $a_{rot}$, and cross-entropy loss for $a_{open}$ and $a_{collide}$. The quaternion distance loss is designed to minimize the difference between the prediction $q_{pred}$ and the ground-truth $q_{GT}$: 
\begin{equation}
    L(q_{pred}, q_{GT}) = 1 - \frac{1}{2}(q_{pred} * q_{GT} + q_{GT} * q_{pred}).
\end{equation}



			



\begin{table*}[htbp]
\vspace*{3mm}
		\caption{
  We present a comparative analysis of our method against baseline methods, reporting task success rates (before slash) and instruction following rates (after slash) on train tasks (above) and novel tasks (bottom). 
  }
  \vspace{-2mm}
    \footnotesize
   \begin{tabularx}{\textwidth} 
    {>{\hsize=0.12\hsize}X
     >{\hsize=0.11\hsize}X
     >{\hsize=0.11\hsize}X
     >{\hsize=0.15\hsize}X
     >{\hsize=0.1\hsize}X
     >{\hsize=0.12\hsize}X
     >{\hsize=0.1\hsize}X
     >{\hsize=0.15\hsize}X
     >{\hsize=0.10\hsize}X}

	\toprule
 
    Train Tasks& close box & close door & open drawer & push button & slide cabinet & turn tap & take umbrella & open bottle\\ 
			
 \midrule
    PERACT & 0.36 / - & 0.72 / - & \textbf{0.44} / - 
    & 0.40 / - & 0.24 / - & 0.76 / - & 0.28 / - & 0.52 / - \\
    
    BC-Z & 0.44 / - & 0.84 / - & 0.00 / - 
    & 0.08 / - & 0.20 / - & 0.00 / - & 0.16 / - & 0.08 / - \\
    
    PERACT + F & 0.36 / - & 0.84 / - & \textbf{0.44} / - 
    & \textbf{0.44} / - & \textbf{0.44} / - & 0.68 / - & 0.48 / -
    & 0.56 / - \\ 
    
    BC-Z + F & 0.48 / 0.37 & \textbf{0.92} / 0.45 & 0.12 / 0.45 
    & 0.20 / 0.44 & 0.08 / 0.33 & 0.00 / 0.23 & 0.28 / 0.58 
    & 0.12 / 0.42 \\ 

    Our w/o Pr & 0.60 / \textbf{0.65} & 0.76 / 0.94 & 0.16 / 0.94 
    & 0.20 / \textbf{0.98} & 0.08 / \textbf{0.90} & 0.40 / 0.85 & 0.24 / 0.99 & 0.60 / 0.65 \\ 

    Ours & \textbf{0.68} / 0.62 & \textbf{0.92} / \textbf{1.00} & 0.24 / \textbf{0.96} 
    & 0.28 / \textbf{0.98} & 0.20 / 0.72 & \textbf{0.80} / \textbf{0.94} & \textbf{0.52} / \textbf{1.00} & \textbf{0.80} / \textbf{0.82} \\ 
    
\midrule

    Novel Tasks& close drawer & close laptop & close microwave & lamp on & open door & open grill & fetch drawer item & Average\\
\midrule
    PERACT & 0.60 / - & 0.40 / - & 0.52 / - 
    & \textbf{0.08} / - & 0.08 / - & 0.28 / - & 0.00 / - 
    & 0.38 / - \\
    
    BC-Z & 0.88 / - & 0.04 / - & 0.32 / - 
    & 0.00 / - & 0.04 / - & 0.00 / - & 0.00 / - 
    & 0.21 / - \\
    
    PERACT + F & 0.68 / - & 0.40 / - & \textbf{0.60} / - 
    & 0.04 / - & 0.08 / - & 0.32 / - & 0.00 / -
    & 0.42 / - \\  
    
    BC-Z + F & 0.88 / 0.44 & 0.16 / 0.29 & 0.36 / 0.48 
    & 0.04 / 0.43 & 0.16 / 0.47 & 0.08 / 0.35 & 0.00 / 0.09 
    & 0.26 / 0.39\\ 

    Our w/o Pr & 0.88 / 0.58 & 0.28 / \textbf{0.59} & 0.28 / 0.74 
    & 0.04 / \textbf{0.69} & \textbf{0.20} / 0.90 & \textbf{0.44} / 0.77 & 0.00 / 0.22
    & 0.34 / 0.76\\ 
    
    Ours & \textbf{0.92} / \textbf{0.86} & \textbf{0.44} / 0.55 & 0.40 / \textbf{0.78} 
    & 0.04 / 0.66 & \textbf{0.20} / \textbf{0.92} & \textbf{0.44} / \textbf{0.91} & 0.00 / \textbf{0.23}
    & \textbf{0.46} / \textbf{0.80}\\ 

\bottomrule
    \end{tabularx}
    
    \label{tab:manipulation}
 \vspace{-2mm}
\end{table*}

\subsection{Prompt module}
The three modalities of vision, language, and manipulation are intricately linked in Vision-and-Language Manipulation tasks.
Concretely, in fine-grained instruction, the verb focuses on action knowledge, which connects the language and manipulation modalities, while the noun focuses on perceptual knowledge, which connects the language and vision modalities. Moreover, similar nouns in different instructions can indicate similar target parts, while similar verbs can indicate similar manipulation actions. 
To better leverage this correspondence between the modalities for learning perceptual and action knowledge, inspired by~\cite{lin2022adapt}, we introduce the Prompt module to explore multi-modal alignment knowledge and help the agent better understand and execute the task.

To begin with, there are two types of prompts in our tasks: perception-prompts, includes a fine-grained instruction and the corresponding affordance map related to the noun, and action-prompts, which combines a fine-grained instruction and the corresponding action denoted by the verb. Since the action cannot be isolated from a specific scene, we add visual observation in action prompt pairs. 
Therefore, we can build two prompt bases for perceiving and action based on the nouns and verbs. The prompt bases can be thought of as a large dictionary, where the key is the important nouns or verbs and the value is the corresponding perception-prompt or action-prompt. 

The Prompt module is designed to learn multi-modal alignment knowledge and consists of two sub-modules: the perception-prompt module and the action-prompt module. 
The perception-prompt module is designed to enhance the Affordance network. During training, the noun-related perception-prompt set (the instruction and the corresponding observation) is retrieved from the pre-built perception-prompt base and fed into the perception-prompt module. The module also uses the pre-trained CLIP to extract the text feature $f_{pr_{noun}}$, and uses the perception-prompt affordance network $PP_{aff}$ to extract the affordance feature $f_{pr_{aff}}$. The $PP_{aff}$ consists of the same PointNet++ model and Affordance network from the bottom part of Fig.~\ref{framework} to predict the affordance map, and it uses the PointNet++ to encode the global feature $f_{pr_{aff}}$ of the affordance map. Finally, the noun-related prompt feature is concatenated with the original input features and fed into the Affordance network.

Similarly, the action-prompt module retrieves the verb-related action-prompt set (the instruction, the corresponding action, and the scene observation) from the pre-built action-prompt base. The module uses the pre-trained CLIP to extract the text feature $f_{pr_{verb}}$, the PointNet++ to extract the observation feature $f_{pr_{vision}}$, and an MLP to extract the action feature $f_{pr_{action}}$. These prompt features are then concatenated with the original input feature and fed into the Actor network. 
To improve the performance of the frozen CLIP model on our task, we add a shared soft-prompt~\cite{liu2021p}, which is a vector of 20-length learnable parameters, to each fine-grained instruction's tokens. The soft-prompt parameters are updated during training, allowing the obtained text features to be better matched to our task without finetuning the CLIP. 
By utilizing the perception and action-prompts, the agent can better leverage cross-modal action knowledge, which is beneficial for guiding correct interaction.

To make each fine-grained language instruction have a closer connection with the corresponding affordance maps and actions, following ~\cite{lin2022adapt}, we introduce the multi-modal alignment loss $\mathcal{L}_{mm}$
Following InfoNCE~\cite{he2020momentum} loss, $\mathcal{L}_{mm}$ aligns the concatenation of the action feature $f_{pr_{action}}$ and visual scene feature $f_{pr_{vision}}$ with the instruction feature $f_{pr_{verb}}$ to encourage multi-modal alignment of paired manipulation and language:
\begin{equation}
   \mathcal{L}_{mm} = -log(\frac{exp(f_{pr_{action\&vision}}\cdot f_{pr_{verb_{+}}}/\tau)}{\sum_{i=0}^{K}exp(f_{pr_{action\&vision}}\cdot f_{pr_{verb_{i}}}/\tau )})
\end{equation}
, where $f_{pr_{action\&vision}}$ donates the concatenation of $f_{pr_{action}}$ and $f_{pr_{vision}}$, $+$ denotes positive pair, $k$ denotes the number of all samples. 

Besides, two fine-grained instructions with the same noun or verb usually lead to similar manipulation, either similar contact regions or similar actions. For example, to "grasp" an umbrella handle or to "grasp" an item in the drawer, the agent should take a similar action: move forward until it reaches the target point and then close the two gripper fingers. Therefore, to encourage the agent to focus on related actions or target object parts, we introduce the consistency loss $\mathcal{L}_{C_{verb}}$ and $\mathcal{L}_{C_{noun}}$.
The $\mathcal{L}_{C_{verb}}$ aims to make the features $f_{pr_{verb1}}$, $f_{pr_{verb2}}$ of two instructions with the same verb closer: 
\begin{equation}
    \mathcal{L}_{C_{verb}} = \left |f_{pr_{verb1}} - f_{pr_{verb2}} \right |.
\end{equation}

The $\mathcal{L}_{C_{noun}}$ aims to make the features of two instructions with the same noun closer, and the loss function is similar to $\mathcal{L}_{C_{verb}}$.
If two fine-grained instructions share the same noun, the two predicted affordances $f_{aff1}$ and $f_{aff2}$ are supposed to highlight similar regions. For example, to grasp the door handle and to grasp the drawer handle, though the two handles probably have different geometric shapes, the two affordance maps should both highlight the two handles. To achieve this, we use the consistency loss $\mathcal{L}_{C_{aff}}$ to make the two affordance map features closer: 
\begin{equation}
    \mathcal{L}_{C_{aff}} = \left |f_{aff1} - f_{aff2} \right |.
\end{equation}

By using these loss functions, we can effectively align the different modalities and improve the agent's understanding and execution of the task.

\section{EXPERIMENTS}

\subsection{Experiment details}
We conduct our experiments in the simulation environment V-REP~\cite{rohmer2013v}, 
interfaced with PyRep~\cite{james2019pyrep}, and use a Franka Panda robotic arm equipped with a parallel gripper. 
In the benchmark, for each task, we offer comprehensive scene information along with demonstration waypoints and 
a script to link the scene objects to the RLBench backend.
It implements task variations, specifying success criteria, and incorporating additional intricate task behaviors. 
Input observations are acquired from four RGB-D cameras strategically positioned: one at the robot's front, one at the left shoulder, one at the right shoulder, and another on the robot's wrist. It is noteworthy that all these cameras are devoid of any noise and boast a resolution of 128 × 128 pixels.

\subsection{Baseline Comparision and Ablation Study}
\paragraph{Evaluation Metric}
We evaluate each multi-task agent independently on 8 training tasks and 7 novel tasks, with each task consisting of 25 evaluation episodes, totaling 375 episodes. We introduce two metrics to quantitatively evaluate the methods. (1) To evaluate the quality of the action proposals, we report the average manipulation success rates per task. During evaluation, the agent continues taking actions until an oracle indicates task completion or until it reaches a maximum of 25 steps.
(2) To evaluate the effectiveness of the Instruction Selection network in understanding the fine-grained instruction sequence based on the current observation, we introduce the metric Instruction Following. During inference, we feed all steps in those 350 testing episodes to the Instruction Selection network and calculate the accuracy of selecting the correct fine-grained instruction.

\paragraph{Baselines}
We compare our approach with four baselines and one ablated version: 
\begin{itemize}
    \item \textbf{I-BC (Image-BC)}, an image-to-action agent similar to BC-Z~\cite{jang2022bc}. Following BC-Z, we use FiLM~\cite{perez2018film} for conditioning with CLIP~\cite{clip} language features. However, the vision encoders take in RGB-D images instead of RGB, and the vision encoder is implemented as a CNN. This baseline only takes in high-level instructions without fine-grained ones.
    \item \textbf{PERACT}, a behavior-cloning agent for multi-task 6-DoF manipulation. It encodes language goals and RGB-D voxel observations with a Perceiver Transformer. This baseline also only takes in high-level instructions without fine-grained ones.
    \item \textbf{I-BC + F}, which is based on I-BC. This baseline takes in additional fine-grained instructions, and we build an Instruction Selection network for it to select the suitable instruction according to the current scene.
    \item \textbf{PERACT + F}, which is based on PerAct. This baseline also takes in additional fine-grained instructions.
    \item \textbf{Our w/o Pr}: an ablated version of our method that removes the Prompt Module.
\end{itemize}

Table.~\ref{tab:manipulation} shows the task success rates (before slashes). We observe that our method outperforms all counterparts. Compared to pure high-level instructions, the fine-grained instructions can improve the manipulation accuracy since they guide the agent to complete the overall tasks step by step. 
Additionally, the ablation study demonstrates that the introduction of the Prompt module further boosts performance by aligning features between different modalities and providing perception and action knowledge.

Table.~\ref{tab:manipulation} also shows the Instruction Following numbers (after slashes). We find that our method can choose the right fine-grained instruction based on the current scene, which demonstrates the effectiveness of our approach in understanding and executing fine-grained instructions.

\subsection{Instruction Generation} \label{Ins Generation}

\begin{figure}[t]
\vspace{3mm}
  \centering
  \includegraphics[scale=0.268]{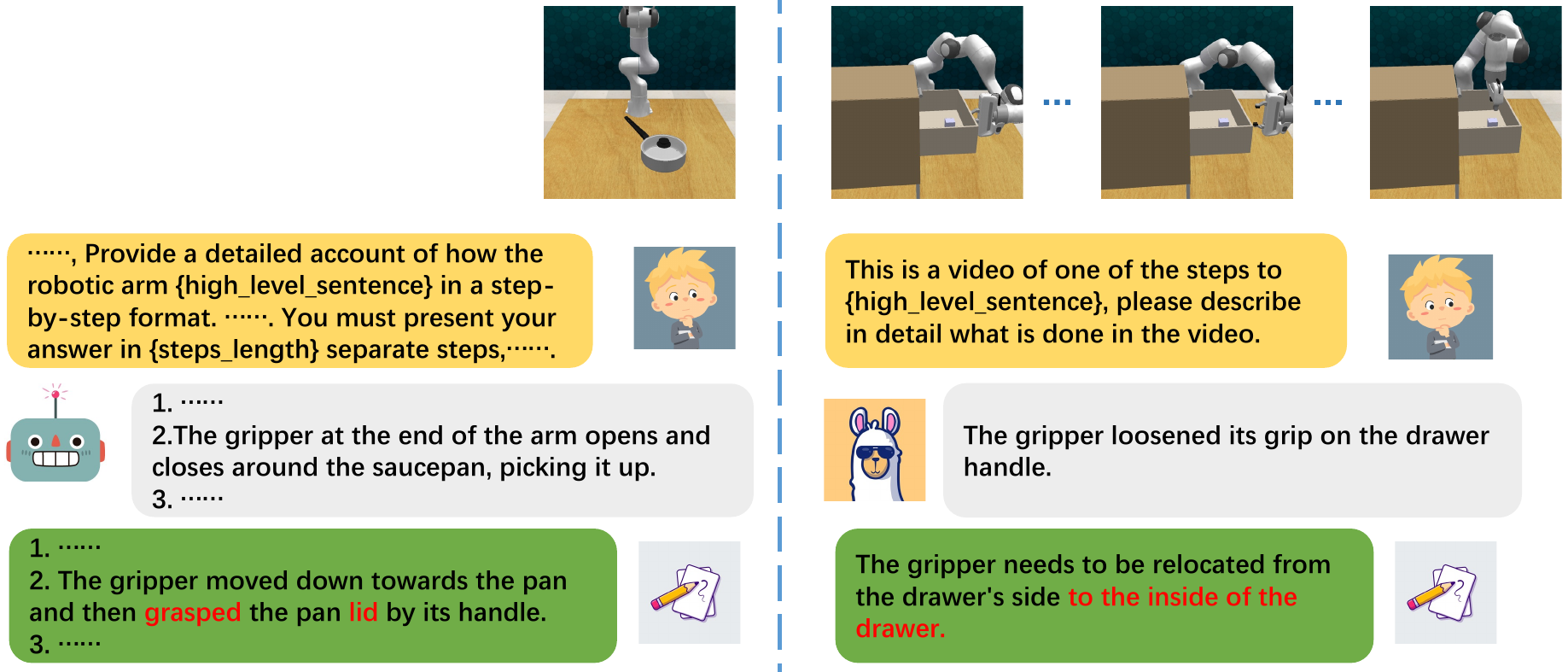}
  \caption{Instruction generating process for two large models (Minigpt4 on the left). The red text in the green box is the important element missed by the large models, ``{high\underline{ }level\underline{ }sentence}'' is the high-level instruction of the current task, and ``{steps\underline{ }length}'' is the total number of steps.}
  \label{fig:InsGene}
\end{figure}

Large language models have revolutionized text annotation by automating labor-intensive and repetitive tasks, significantly reducing human effort. This innovation has accelerated the traditionally time-consuming process, making it more efficient. These models are also highly adaptable to specific tasks without requiring retraining, showcasing their versatility across various domains. Moreover, they excel in minimizing annotation errors, thereby elevating data quality and boosting the performance of machine learning and natural language processing applications.

\paragraph{Models}
To evaluate the effectiveness of large language models in offering precise instructions for our NrVLM dataset, we conduct experiments employing two distinct models: Minigpt4~\cite{zhu2023minigpt} and Video-LLaMA~\cite{zhang2023video}. Specifically, Minigpt4 is utilized for tasks involving image-based question and answer tasks. For this, we select an RGB image captured from a frontal perspective of each manipulation episode, with the corresponding question template depicted in Fig.~\ref{fig:InsGene}.
On the other hand, the Video-LLaMA model adopts video-based question and answer tasks. Here, we input the video corresponding to each manipulation step to derive answers based on the question templates showcased in Fig.~\ref{fig:InsGene}. To enhance the Video-LLaMA model's suitability for manipulation tasks, we fine-tune it using the training set tasks from our benchmark.

\begin{table}[t]
\vspace{3mm}
\caption{Quantitative results of the language instructions generated by large language models, demonstrating the value of human-annotated fine-grained instructions.}
\label{tab:ins}
\centering
\begin{adjustbox}{width=0.95\linewidth}
\begin{tabular}{ccc}
\hline
Method & Bleu score & Rouge score \\
\hline
Minigpt4 & 11.94 & 30.48 \\
\hline
Video-LLaMA & 7.11 & 36.74 \\
\hline
Video-LLaMA (fine-tune) & 15.07 & 34.74 \\
\hline
\end{tabular}
\end{adjustbox}
\end{table}

\paragraph{Metrics}
We adopt two primary metrics in natural language processing, namely Bleu (Bilingual Evaluation Understudy)~\cite{brown1990statistical} and Rouge (Recall-Oriented Understudy for Gisting Evaluation)~\cite{lin2004rouge}.

The Bleu score plays a pivotal role in assessing the accuracy of machine-generated translations in natural language processing tasks, particularly in machine translation. It serves as a quantitative measure to gauge the degree of similarity between machine-generated text and human reference translations. The Bleu score provides a precise and objective assessment of translation accuracy by quantifying the alignment between machine-generated translations and human references.
The Rouge score plays a crucial role by quantifying the degree of overlap between the content of machine-generated summaries and that of human reference summaries. This overlap is measured through the identification of shared n-grams (sequences of n words) between the machine-generated and reference summaries, providing insights into the extent to which key information has been successfully captured and conveyed.

The Rouge score assesses the ability of summarization systems to capture and retain essential information from the source text, ensuring that the generated summaries adequately represent the key points and main ideas contained therein.
In our experiments, manually labeled language instructions serve as ground truth, and these two metrics are employed to assess the quality of the results generated by the two large language models.

\paragraph{Analysis}
The Bleu and Rouge scores are both in the range of 0 to 100, and the results in Table.\ref{tab:ins} indicate that both Minigpt4 and Video-LLaMA's generated results perform very poorly in terms of accuracy and recall. After fine-tuning on the manipulation tasks in our training set, Video-LLaMA's generated results can achieve a little gain on Bleu compared to the original Video-LLaMA, but the overall quality of the generation is still very low.
These outcomes clearly emphasize that although the large language model demonstrates a degree of understanding when presented with input images or videos, its accuracy notably lags behind in generating manipulation outputs, let alone serving as instructions for guiding agent movement. 
As a result, the manually curated fine-grained natural instructions in our NrVLM benchmark hold significant value and necessity, particularly in tasks that involve intricate planning and precise manipulation.

\section{CONCLUSIONS}

In this paper, we introduce the NrVLM benchmark, offering detailed natural instructions to assist agents in executing complex tasks sequentially. Alongside this benchmark, we present a framework for instructions following and establishing a manipulation-aware multi-modality alignment, enhancing precise manipulation prediction. 

\bibliographystyle{plain}
\bibliography{root}

\end{document}